\def\year{2021}\relax
\documentclass[letterpaper]{article} 
\usepackage{aaai21}  
\usepackage{times}  
\usepackage{helvet} 
\usepackage{courier}  
\usepackage[hyphens]{url}  
\usepackage{graphicx} 
\usepackage{natbib}  
\usepackage{caption} 
\usepackage{epsfig}
\usepackage{graphicx}
\usepackage{amsmath}
\usepackage{amssymb}

\usepackage{enumerate}
\usepackage{booktabs}
\usepackage{bbm}
\usepackage{dsfont}
\usepackage{enumerate}
\usepackage{amsmath,amssymb} 
\usepackage{multirow} 
\usepackage{amsmath} 
\usepackage{amssymb}
\usepackage{commath} 
\usepackage{enumitem} 
\usepackage{fancyhdr} 
\DeclareMathOperator*{\argmax}{argmax}

\title{Language Guided Networks for Cross-modal Moment Retrieval}
\author{}
\affiliations{}
\begin{document}

\maketitle

\begin{abstract}
We address the challenging task of cross-modal moment retrieval, which aims to localize a temporal segment from an untrimmed video described by a natural language query.
It poses great challenges over the proper semantic alignment between vision and linguistic domains.
Existing methods independently extract the features of videos and sentences and purely utilize the sentence embedding in the multi-modal fusion stage, which do not make full use of the potential of language.
In this paper, we present Language Guided Networks (LGN), a new framework that leverages the sentence embedding to guide the whole process of moment retrieval.
In the first feature extraction stage, we propose to jointly learn visual and language features to capture the powerful visual information which can cover the complex semantics in the sentence query.
Specifically, the early modulation unit is designed to modulate the visual feature extractor's feature maps by a linguistic embedding.
Then we adopt a multi-modal fusion module in the second fusion stage.
Finally, to get a precise localizer, the sentence information is utilized to guide the process of predicting temporal positions.
Specifically, the late guidance module is developed to linearly transform the output of localization networks via the channel attention mechanism.
The experimental results on two popular datasets demonstrate the superior performance of our proposed method on moment retrieval
(improving by 5.8\% in terms of Rank1@IoU0.5 on Charades-STA and 5.2\% on TACoS).
The source code for the complete system will be publicly available.
\end{abstract}
\section{Introduction}

\label{intro}
Joint modeling of videos and language, such as video captioning \cite{pan2017video}, 
video question answering \cite{fan2019heterogeneous}, 
and video generation conditioned on captions \cite{pan2017create}, 
has attracted considerable attention in artificial intelligence community due to pervasive applications.
In this paper, we focus on the cross-modal moment retrieval task \cite{anne2017localizing,liu2018attentive}, a natural testbed for integrating sentences and videos,
which aims to temporally localize activities from unconstrained videos via language ~\cite{gao2017tall}.

\begin{figure*}[t]
\begin{minipage}[b]{0.96\linewidth}
  \centering
  \centerline{\includegraphics[width=18cm,height=3.25cm]{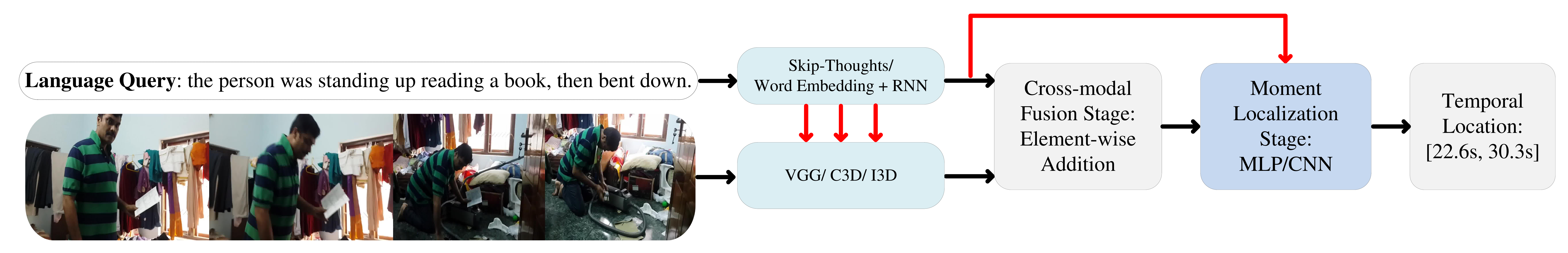}}
\end{minipage}
\caption{An overview of our moment retrieval pipeline.
As indicated by the red arrows, 
we propose to modulate the visual feature processing conditioned on sentence. 
Besides, sentence embedding is further utilized to guide the localization networks to determine the start and end time of the moments.}
\label{fig:01}
\end{figure*}

As shown in Figure \ref{fig:01}, given a natural language query and an untrimmed video,
the goal is to localize the start and end time of the moments described by the given sentence query.
Compared with traditional video retrieval tasks, the queries of cross-modal moment retrieval are often
complex and diverse, which consist of many semantic concepts and can be arbitrary natural descriptions.
Take the sentence ``the person was standing up reading a book,
then bent down'' in Figure \ref{fig:01} as an example,
besides the actor ``person'' and object ``book'', the activity contains several actions,
namely ``standing up'', ``reading'' and ``bent down''.
Therefore, this task requires deep multimodal comprehension which can tightly integrate vision and language features.

Current approaches for moment retrieval tasks mainly consist of the following stages as illustrated in Figure \ref{fig:01}:
feature extraction stage for encoding the videos and sentences into embedding,
cross-modal fusion stage for integrating the information from both modalities,
and moment localization stage for identifying the start and end time of the described activity inside the video.

More specifically, these methods usually first extract powerful video features using 3D convolutional neural network (CNN) (e.g. the C3D \cite{tran2015learning}) or
an ImageNet pre-trained CNN (e.g. VGG16 \cite{Simonyan15}),
and obtain a sentence embedding via a LSTM over word embeddings \cite{pennington2014glove}
or an off-the-shelf sentence encoder (e.g. Skip-thought \cite{kiros2015skip}).
Then sentence embedding and video features are fused in the cross-modal feature fusion stage, such as the element-wise addition used in ~\cite{gao2017tall} and vector concatenation used in ~\cite{wang2019language}.
Finally, these methods predict the temporal position in the localization stage,
which is usually composed of MLP \cite{liu2018attentive}, temporal convolutional networks \cite{yuan2019semantic} or 2D-CNN \cite{2DTAN}.

Despite the promising performance,
the majority of existing methods do not integrate the video features and language tightly.
We believe there are two reasons for this.
First of all, the recent literature independently processes each modality in the feature extraction stage.
Pre-trained CNNs are good choices to extract feature since it can prevent over-fitting.
However, these extractors are unlikely to capture the complex semantics of sentence query
because they are usually trained on a limited set of actions or objects described at the word level.
Second, some state-of-the-art systems ignore the sentence embedding in the localizer stage.
Although these methods combine the language and video in the fusion stage,
it is far from enough for moment localization due to
its high requirement for the proper semantic alignment between the vision and linguistic domain.

To solve the above challenges, unlike previous work that only utilizes language in the fusion stage,
we propose the Language Guided Networks (LGN) that leverage sentence embedding to guide the whole process of moment retrieval.
As shown in the red arrows in Figure \ref{fig:01}, in the first feature extraction stage,
we modulate the visual processing of the pre-trained models (e.g., VGG) conditioned by the given sentence embedding,
capturing the more diverse visual concepts to fit the complex semantics of the sentence query.
Specifically, we develop the early modulation unit to update the feature maps of CNN.
In the cross-modal feature fusion stage, we combine the vision and text features
via the widely used Hadamard product.
Finally, we utilize the sentence embedding to guide the localizer
to predict the temporal positions of the moments,
which can further integrate the vision and language feature.
Specifically, the sentence embedding is linearly transformed to generate the attention weight for each channel of feature maps.
Experiments on two challenging benchmarks demonstrate that
the proposed method significantly boosts the performance
and outperforms prior works.
In particular, the proposed method can achieve 48.1\% and 30.5\% in Rank1@IoU5 on the Charades-STA and TACoS, respectively.

To summarize, our contributions are as follows:
 \begin{itemize}
    \item To the best of our knowledge, we are the first to adopt the sentence embedding to guide the whole process of moment retrieval.
    \item We design the early modulation unit to modulate the visual processing by sentences in the feature extraction stage, which can generate powerful visual features.
    \item In the moment localization stage, we develop the late guidance module to update the outputs of the localization networks, which considerably improves the retrieval performance.
    \item We conduct extensive experiments on two datasets to demonstrate that the proposed method outperforms
the state-of-the-art methods by a substantial margin. As a side contribution, we will make codes publicly available if accepted.
\end{itemize}

\section{Related Work}
\label{relate}
Moment localization with natural language is a new vision-language task introduced in \cite{anne2017localizing,gao2017tall}.
According to their key characteristics, we group the state-of-the-art approaches into three categories:
two-stage models, one-stage methods, and reinforcement learning-based approaches.

Some previous methods \cite{anne2017localizing,gao2017tall} train two-stage models which adopt the sliding windows to generate moment candidates and then match them with the sentences. 
Specifically, Hendricks et al. \cite{anne2017localizing} combine temporal endpoint features with video features.
Gao et al. \cite{gao2017tall} design a temporal regression network to align the temporal position between
the candidate moment and the target clip.
Liu et al. \cite{liu2018attentive} develop a memory attention mechanism to emphasize the visual features described in the sentence.
Ge et al. \cite{ge2019mac} extend the previous work \cite{gao2017tall}
by mining verb-object pairs in the sentence to further facilitate the localization.
Similarly, Jiang et al. \cite{jiang2019cross} investigate the objects in the language to retrieve moments
via a spatial and language-temporal attention model.
These approaches can achieve competitive results but perform inefficiently.

Yuan et al. \cite{yuan2019find} propose a single-shot attention based method that directly predicts the temporal coordinates of the video segment that described by the sentence query.
Chen et al. \cite{chen2018temporally} develop a temporal groundnet
to evolve fine-grained frame-by-word interactions across video-text modalities.
To better capture the sentence-related video contents over time,
a novel Semantic Conditioned Dynamic Modulation (SCDM) mechanism is proposed in \cite{yuan2019semantic}
to modulate the temporal convolution operations conditioned by the sentence semantics.
Recently, a Dense Regression Network (DRN) is proposed in \cite{zeng2020dense} to increase the number
of positive training samples, which gains outstanding performance of moment retrieval.
These models ~\cite{2DTAN,yuan2019semantic,zeng2020dense,yuan2019find} are effective and efficient since they do not need to exhaustively enumerate candidates.

Inspired by the success of applying reinforcement learning into vision tasks,
He et al. \cite{he2019read} first adopt the reinforcement learning paradigm for moment retrieval
and learn an agent to read the description, to watch the video, and then to move the temporal grounding boundaries.
Wang et al. \cite{wang2019language} propose a RNN based reinforcement learning framework
that matches the given sentence with video clips in a matching-based manner.
Wu et al. \cite{wu2020tree} further extend the previous work \cite{he2019read}
via a tree-structured policy which aims to reason the robust primitive actions in the language.
This kind of method formulates the task as a sequential decision-making problem, rather than treating it as a regression or ranking problem.
Although these methods predict the temporal moment in a single-stage manner, we list them separately for clarity.

Xu et al. \cite{xu2019multilevel} leverage the sentence to filter out irrelevant clips in the early stage, which is one of the closest work to ours.
Compared with this work, our approach integrates cross-modal features more tightly
since it not only adopts the sentence embedding to modulate the visual feature extraction of CNNs,
but also further utilizes sentences to guide the process of predicting temporal positions.

\section{The Proposed Framework}
\label{thepr}
In this section, we first describe the overall framework of LGN. 
We then introduce the basic formation of moment retrieval with natural language.
Finally, we introduce the main stages of our method:
feature extraction stage, the multi-modal fusion stage, and moment localization stage.

\subsection{Overall Framework}
\begin{figure*}[t]
\begin{minipage}[b]{0.96\linewidth}
  \centering
  \centerline{\includegraphics[width=16cm,height=4.8cm]{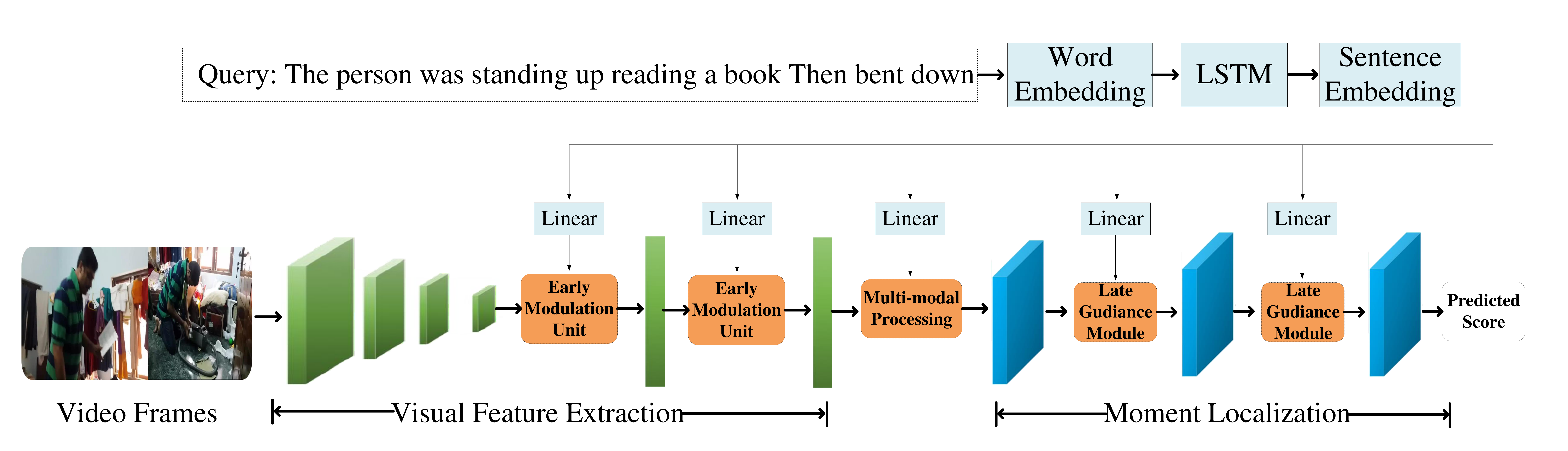}}
\end{minipage}
\caption{The proposed overall framework for cross-modal moment retrieval.
We propose to leverage the sentence embedding to guide the whole process for retrieving moments.
First, we design the early modulation units to modulate the visual feature processing conditioned on sentence.
Besides, the late guidance module is developed to utilize the language to update the output of localizer.}
\label{fig:02}
\end{figure*}

Figure \ref{fig:02} shows the framework of the proposed approach.
We leverage the sentence embedding to guide the whole process of moment retrieval.
First, we manipulate the feature maps of a target CNN conditioned on
the sentence semantics via the early modulation unit in the feature extraction stage.
Second, visual representation and sentence embedding are further integrated
in the multi-modal fusion stage.
Third, the sentence embedding is adopted to guide the process of retrieval
through the late guidance module in the moment localization stage.

\subsection{Problem Definition}
Given an untrimmed video $V$ and a sentence $S$ as a query,
our task aims to retrieve a moment candidate $M$ in $V$ that matches the $S$ best.
More specifically, we denote an untrimmed video as $V$ = ${\{V_i}\}$ $^{l_v} _{i=1}$ , where ${V_i}$ represents a moment proposal and ${l_v}$ is the total number of moment candidates.
Besides, we denote the query sentence as $S$ = ${\{s_i}\}$ $^{l_s} _{i=1}$,
where $s_i$ represents a word among the sentence, and ${l_s}$ is the total number of words.
Each training sample is annotated with the start and endpoint of the sentence.
We adopt the intersection over union between ground-truth and moment candidates to train networks.
At test time, given V and S, the retrieval problem is formulated as
 \begin{equation}
   M = \mathop{\argmax}_{V_i \in V}{\sigma(V_i,S)},
 \label{equ:01}
 \end{equation}
where $\sigma$ is the trained network that predicts the IOU score between $V_i$ and the target moment.
The target moment best matches S.
Note that we formulate the moment retrieval as a classification problem since the experiments in 2D-TAN \cite{2DTAN} demonstrate that selecting the right moment candidates can achieve rather ideal performance even without regression.

\subsection{Feature Extraction Stage}
\textbf{Sentence Representation.}
A sentence encoder is a function $F_s(s_i)$ that maps all words $s_i$ into a sentence-level embedding.
As shown in Figure \ref{fig:02}, we first generate the embedding of each word $w_i$ $\in$ $R^{d_s}$ by the GloVe word2vec model \cite{pennington2014glove}, where ${d_s}$ is the length of embedding.
Then, the LSTM network is adopted to encode each word embedding sequentially. 
Finally, we linearly transform the last hidden state of LSTM and
take it as the feature representation of the input sentence,
which we denote as $f_s$ $\in$ $R^{d_s}$ for the rest of this paper.

\textbf{Early Modulation Unit.}
In this section, we propose a novel method to modulate visual processing by a linguistic input.
In previous methods, the video features are extracted through off-the-shelf visual encoders,
such as the C3D ~\cite{tran2015learning} and an ImageNet pre-trained VGG \cite{Simonyan15}. 
As a result, visual features extracted from these methods cannot
cover the complex semantic of sentence query
since they usually contain the word level information, such as the nouns of ImageNet.
These findings suggest that independently extracting visual and language features might be suboptimal.
Thus, we propose to couple sentence embedding and visual processing via the early modulation unit.

The early modulation unit aims to influence
the feature of a neural network by applying a shift operation.
More specifically, we drive the neural networks to cover the semantic of sentences
via the integration of both media.
We couple the feature maps of the networks and the sentence embedding via the Schur product.
Formally, the early modulation unit can be defined as
 \begin{equation}
   F = (W^Mf_s) \circ (f_v),
   \label{equ:02}
 \end{equation}
 where $W^M$ is the learned parameters of multilayer perceptrons (MLP), $f_s$ is the sentence embedding,
 $\circ$ represents the Schur product operation, and $f_v$ is the feature map of neural networks.
 The goal of $W^M$ is to linearly transform $f_s$ into
 the same dimension with $f_v$. We share the same $W^{M}$ in the feature extraction stage for fewer parameters.

Through experiments, we found that adding the early modulation units after convolutional layers
yields unsatisfactory results,
which is different from \cite{perez2018film} that inserts the modulation unit into ResBlock.
The reason may be that convolutional blocks usually capture the low-level features.
Thus, early modulation units are equipped at the fully connected layer which
always contains more semantic information.

Besides, we follow the 2D-TAN \cite{2DTAN} to organize
the moment proposals as a two-dimensional temporal map,
where one dimension indicates the start time of a moment
and the other suggests the endpoint.
This approach achieves outstanding results because of
the powerful ability to learn adjacent temporal relations.
We also demonstrate the superior performance of the early modulation unit
based on 2D-TAN ~\cite{2DTAN} in experiments section. 

\subsection{Multi-modal Fusion Stage}
The inputs of the multi-modal fusion stage are sentence embedding and the modulated visual representation,
which have the same dimension.
Specifically, we embed these two cross-domain features into a unified subspace by fully-connected
layers, and then fuse them through Hadamard product and $L2$ normalization.

\subsection{Moment Localization Stage}
\textbf{Late Guidance Module.}
After obtaining the multi-modal representation,
we further guide the process of moment localization through the sentence embedding.
Following 2D-TAN \cite{2DTAN}, we adopt the 2D-CNN as the backbone of moment localization.
More specifically, we first shift the length of sentence embedding
into the channel numbers of the feature map of 2D-CNN.
Then, we employ the sentence embedding as channel attention for the outputs of 2D-CNN.
Specifically, we update each channel by multiplying the corresponding scalar $\alpha_{i}$ from the sentence embedding.
Finally, a normalization operation is conducted to avoid over-fitting.

This module is achieved by the following equation:
 \begin{equation}
   C_{i}^{'} = \frac {\alpha_{i} \times C_{i}}  {\norm {\alpha_{i} \times C_{i}}_2 },   \alpha = W^{M'}f_s,
   \label{equ:03}
 \end{equation}
where the $\alpha_{i}$ is the $i_{th}$ of learned scalars $\alpha$,
$C_{i}$ is $i_{th}$ channel of the feature map of 2D-CNN,
$f_s$ is sentence embedding, and $W^{M'}$ is the learned parameters of MLP.

The late guidance module can be regarded as a channel attention mechanism
designed for the outputs of localization networks.
Besides, various $\alpha_{i}$ can transform individual feature maps in a variety of ways.
More importantly, ${\alpha}$ can offer valuable guidance to the moment retrieval
since it is derived from the sentence embedding.
Moreover, as we share the same $W^{M'}$ in the moment localization stage,
the late guidance module is a computationally efficient method and only involves a handful of parameters.

In particular, late guidance module can be plugged easily into other moment localization architectures,
such as the temporal convolutional networks adopted in \cite{yuan2019semantic}
and 1-dimensional convolutional networks used in \cite{zhang2019man}.

\textbf{Loss Function.}
During the training, we adopt a normalized IoU value as the supervision label,
rather than a hard binary score.
More specifically, we compute the temporal IoU for each moment proposal and the target clip.
Then IoU score $o_i$ is then shifted by min-max normalization
with two  hyper-parameters $t_{min}$ and $t_{max}$ to generate the supervision signal $y_i$.
Formally,
\begin{equation}
 y_i =\left\{
\begin{array}{rcl}
 0      &      & {o_i     \leq     t_{min} },\\
\frac{o_i - t_{min}}{t_{max} - t_{min}}     &      & {t_{min}  <  o_i < t_{max}},\\
 1    &      & {o_i \geq t_{max}.}
\end{array} \right.
\label{equ:04}
\end{equation}
Then, our framework is trained via a binary cross entropy loss as
\begin{equation}
    Loss(y_i,p_i)= -\frac{1}{N}\sum_{i=1}^{N}{y_i\log p_i + (1-y_i)\log (1-p_i)},
\label{equ:05}
\end{equation}
where $p_i$ is the predicted score of a moment candidate
and $N$ is the total number of proposals.

\section{Experiments}
\label{exper}
To evaluate our approach, we conduct experiments on two popular and challenging benchmarks:
Charades-STA \cite{gao2017tall} and TACoS \cite{regneri2013grounding}.
In this section, we first describe both datasets, evaluation metrics, and implementation details.
Then, we compare the performance of our method with the state-of-the-art models.
Finally, ablation studies are introduced to further demonstrate the effectiveness of our method. 

\subsection{Datasets}
\textbf{Charades-STA.} Gao et al. \cite{gao2017tall} establish the Charades-STA by annotating the temporal position and sentence description of Charades \cite{sigurdsson2016hollywood},
which is originally built for daily activity recognition and localization.
In Charades-STA, there are 12,408 moment-sentence pairs in the training set,
and 3,720 moment-sentence pairs in the test set.
Since the Charades-STA dataset only releases the moment-sentence description file,
we download the frames from the official website \footnote{https://prior.allenai.org/projects/charades}
to extract the features of moment proposals.

\textbf{TACoS.} Regneri et al. \cite{regneri2013grounding} build the TACoS by adding the natural language descriptions of MPII Composites dataset \cite{rohrbach2012script}, which contains diverse fine-grained activities in the kitchen.
They also align each sentence to its corresponding video moments
to make TACoS suitable for moment retrieval.
In total, there are 17,344 moment-sentence pairs.
Following the same train/val/test split strategy as ~\cite{gao2017tall,2DTAN},
We split it in 50\% for training, 25\% for validation, and 25\% for test.

To further analyze the characteristics of both benchmarks,
we list the average number of words per sentence,
the average number of moments per video, and the average duration of the moments
in Table \ref{tab:00}.
Although the videos in TACoS is less than that of Charades-STA (127 versus 6,672),
TACoS has more moment-sentence pairs 
since TACoS has more temporally annotated video segments with queries per video.
It seems that the TACoS is more challenging due to more words and the same scene in the kitchen.

\renewcommand\arraystretch{1.2}
\begin{table}[h]\footnotesize
\begin{center}
\newcommand{\tabincell}[2]{\begin{tabular}{@{}#1@{}}#2\end{tabular}}
\centering \caption{Statistics of the Charades-STA and TACoS datasets.}
\label{tab:00}
\vspace{1mm}
\begin{tabular}{ccccccc}
  \toprule
         Dataset    &\tabincell{c}{\#Words}  &\tabincell{c}{\#Moment}  &\tabincell{c}{\#Duration}   &\tabincell{c}{Domain}\\
  \midrule
      Charades-STA         &6.2 &2.3 &8.2   &Homes\\
  \midrule
      TACoS         &8.5 &136.6 &27.9 &Kitchen \\
  \bottomrule
\end{tabular}
\end{center}
\end{table}
\subsection{Evaluation Metrics}
Following previous work \cite{gao2017tall,2DTAN},
we adopt Rank n@ tIoU = m metric to evaluate our method.
It is defined as the percentage of sentence queries having at least one correct moment retrieval
in the top $n$ retrieved clips.
A retrieved clip is correct when its tIoU (temporal Intersection Over Union)
with the target clip is larger than $m$.
For fair comparison, we adopt the same $n$ and $m$ with previous work ~\cite{2DTAN}.
Specifically, we use n $\in$ \{1, 5\} with m $\in$ \{0.5, 0.7\} for Charades-STA dataset,
and n $\in$ \{1, 5\} with m $\in$ \{0.1, 0.3, 0.5\} for TACoS dataset.

\subsection{Implementation Details}
To compare fairly with the state-of-the-art approaches \cite{2DTAN,zhang2019man,yuan2019semantic},
we utilize the VGG \cite{Simonyan15} model from the Charades official website. \footnote{https://github.com/gsig/charades-algorithms/tree/master/pytorch}
for Charades-STA dataset, and C3D \cite{tran2015learning} for TACoS dataset.
A recurrent LSTM over the word embedding \cite{pennington2014glove} is adopted for encoding sentences.
We mostly set the same hyper-parameters as 2D-TAN \cite{2DTAN} for a fair comparison. 
We adopt the common pooling operation to extract moment features from clip features.
The dimensions of $f_v$, $f_s$, $w_i$ are $4096$, $512$ and $300$ respectively.
The $t_{min}$ and $t_{max}$ are $0$ and $0.5$ respectively.
We adopt two early modulation units and six late guidance modules.
In the early modulation unit, we add two dropout layers after the fully-connected layers with a ratio of $0.75$.
The networks are optimized by Adam optimizer
and the whole system is implemented via the PyTorch toolkit \cite{paszke2017automatic}.
All experiments are conducted on a single server with $4$ GeForce GTX TITAN XP cards.

\subsection{Comparison with The-state-of-the-art Methods}
\textbf{Quantitative results.} We compare the performance of our method with the following state-of-the-art methods:
\begin{enumerate}
    \item Two-stage approaches: this kind of method always adopts the sliding windows to generate multi-scale moment candidates and then match them with the sentences. A lots of sliding windows based two-stage approaches are proposed, including MCN \cite{anne2017localizing}, CTRL \cite{gao2017tall}, MAC \cite{ge2019mac}, SLTA ~\cite{jiang2019cross}. 
    \item One-stage single-shot methods: this kind of method always combines both video and language information in a single-shot structure to directly predict the matching probabilities between moment candidates and sentences. These methods are efficient since they only require to process the video in one pass. The representative work includes ABLR \cite{yuan2019find}, SCDM \cite{yuan2019semantic} and DRN \cite{zeng2020dense}.
    \item Reinforcement learning-based methods: this kind of method introduce the reinforcement learning paradigm for the moment retrieval. Some representative approaches are recently proposed, including RWM \cite{he2019read}, SM-RL \cite{wang2019language}, TSP-PRL \cite{wu2020tree}.
\end{enumerate}

\renewcommand\arraystretch{1.06}
\begin{table*}[t]\footnotesize
\begin{center}
\newcommand{\tabincell}[2]{\begin{tabular}{@{}#1@{}}#2\end{tabular}}
\centering \caption{Comparison of state-of-the-art methods on TACoS dataset.}
\label{tab:02}
\vspace{1mm}
\begin{tabular}{cccccccc}
  \toprule
       &Methods    &\tabincell{c}{Rank1@\\IOU0.1(\%)}  &\tabincell{c}{Rank1@\\IOU0.3(\%)}
          &\tabincell{c}{Rank1@\\IOU0.5(\%)}          &\tabincell{c}{Rank5@\\IOU0.1(\%)}  &\tabincell{c}{Rank5@\\IOU0.3(\%)} &\tabincell{c}{Rank5@\\IOU0.5(\%)}    \\
  \midrule
    ~ &SM-RL \cite{wang2019language}       &26.51 &20.25 &15.95 &50.01 &38.47 &27.84\\
     ~ &TripNet \cite{hahn2019tripping}       &--    &23.95 &19.17 &--    &--    &--\\
  \midrule 
  &MCN \cite{anne2017localizing}         &14.42 &-- &5.58 &37.35 &-- &10.33\\
  &ROLE \cite{liu2018cross}           &20.37 &15.38 &9.94 &45.45 &31.17 &20.13\\
  &SLTA \cite{jiang2019cross}           &23.13 &17.07 &11.92 &46.52 &32.90 &20.86\\
  &CTRL \cite{gao2017tall}              &24.32 &18.32 &13.30 &48.73 &36.69 &25.42\\
  &ACRN \cite{liu2018attentive}         &24.22 &19.52 &14.62 &47.42 &34.97 &24.88\\
  &QSPN \cite{xu2019multilevel}          &25.31 &20.15 &15.23 &53.21 &36.72 &25.30\\
  &VAL \cite{song2018val}                &25.74 &19.76 &14.74 &51.87 &38.55 &26.52\\
  &MCF \cite{wu2018multi}                &25.84 &18.64 &12.53 &52.96 &37.13 &24.73 \\
  &SAP \cite{chen2019semantic}          &31.15 &-- &18.24 &53.51 &-- &28.11\\
  &ACL-K \cite{ge2019mac}                &31.64 &24.17 &20.01 &57.85 &42.15 &30.66\\
  &CMIN \cite{zhang2019cross}            &32.48 &24.64 &18.05 &62.13 &38.46 &27.02 \\
  \midrule
  &ABLR \cite{yuan2019find}              &34.70 &19.50 &9.40 &-- &-- &--\\
  &TGN \cite{chen2018temporally}        &41.87 &21.77 &18.9 &53.40 &39.06 &31.02 \\
  &SCDM \cite{yuan2019semantic}           &-- &26.11 &21.17 &-- &40.16 &32.18 \\
  &DRN \cite{zeng2020dense}              &-- &-- &23.17 &-- &-- &33.36 \\
  &2D-TAN \cite{2DTAN}                   &47.59 &37.29 &25.32 &70.31 &57.81 &45.04\\
  &Ours      &\textbf{52.46}   &\textbf{41.71}  &\textbf{30.57} &\textbf{76.86} &\textbf{63.06} &\textbf{50.76} \\
  \bottomrule
\end{tabular}
\end{center}
\end{table*}

\renewcommand\arraystretch{1.0}
\begin{table}[t]\footnotesize
\begin{center}
\newcommand{\tabincell}[2]{\begin{tabular}{@{}#1@{}}#2\end{tabular}}
\centering \caption{Comparison of state-of-the-art methods on Charades-STA dataset.}
\label{tab:01}
\vspace{1mm}
\begin{tabular}{cccccc}
  \toprule
            &Methods   &\tabincell{c}{Rank1@\\IOU0.5}  &\tabincell{c}{Rank1@\\IOU0.7}          &\tabincell{c}{Rank5@\\IOU0.5}  &\tabincell{c}{Rank5@\\IOU0.7}\\
  \midrule 
      &SM-RL          &24.36 &11.17 &61.25 &32.08\\
      &RWM                  &34.12 &13.74 &- &- \\
      &TripNet       &36.61 &14.50 &- &- \\
      &TSP-PRL             &45.30 &24.73 &- &- \\
  \midrule 
      &MCN          &17.46 &8.01 &48.22 &26.73\\
      &ACRN          &20.26 &7.64 &71.99 &27.79\\
      &ROLE            &21.74 &7.82 &70.37 &30.06\\
      &SLTA             &22.81 &8.25 &72.39 &31.46\\
      &VAL                 &23.12 &9.16 &61.26 &27.98\\
      &CTRL               &23.63 &8.89 &58.92 &29.52\\
      &SAP           &27.42 &13.36 &66.37 &38.15\\ 
      &ACL-K                 &30.48 &12.20 &64.84 &35.13\\
      &QSPN           &35.60 &15.80 &79.40 &45.40\\
      &MAN                &41.24 &20.54 &83.21 &51.85\\
  \midrule 
      &ABLR               &24.36 &9.01 &- &- \\
      &2D-TAN                  &40.94 &22.85 &83.84 &50.35 \\
      &DRN              &42.90 &23.68 &\textbf{87.80} &\textbf{54.87}\\
      &Ours      &\textbf{48.15}   &\textbf{26.67}  &86.80 &53.01 \\
  \bottomrule
\end{tabular}
\end{center}
\end{table}


The comparison between our method and state-of-the-art methods are presented in Table \ref{tab:02} for TACoS ~\cite{regneri2013grounding} dataset and Table \ref{tab:01} for Charades-STA \cite{gao2017tall} dataset.
From both tables, we can see that our approach outperforms all of the baselines in terms of Rank1@IoU0.5.
For clarity, we mark the best results in bold font. We have the following observations.

First, the typical two-stage methods, such as MCN \cite{anne2017localizing}, CTRL \cite{gao2017tall},
ACRN \cite{liu2018attentive} all extract the visual feature and language vector independently,
which causes the failure of visual features to cover the comprehensive semantic in sentences.
In contrast, our approach makes the visual model capture the diverse concepts by
modulating the visual feature extraction conditioned by the sentence embedding.
For example, our method improves Rank1@IoU0.5 by 12.25\% compared with the
previous representative two-stage method \cite{xu2019multilevel} on the Charades-STA.

Second, we observe that single-shot models usually obtain better performance than two-stage approaches.
However, most of these methods combine vision and text features once in the multimodal fusion stage.
Some work \cite{zhang2019man,xu2019multilevel} integrates cross-modal features twice.
For example, \cite{xu2019multilevel} adopts the linguistic input to filter out background moments in the early stage.
In contrast, our method fuses the multi-modal features more tightly because our method
not only adopts the sentence semantics to modulate the visual feature
extraction of CNNs but also utilizes sentences to guide
the process of predicting temporal positions.
Compared with the previous best one-stage methods \cite{zeng2020dense},
our method surpasses it by 5.25\% on the Charades-STA in terms of Rank1@IoU0.5 using the same VGG16 model.

Third, although outstanding success is achieved by reinforcement learning,
our method still shows superior performance over all reinforcement learning-based methods.
For example, the recent reinforcement learning-based method \cite{wu2020tree} designs a tree-structured policy
and achieves promising performance with the two-stream features.
However, it is still inferior to ours using the RGB features.

We further evaluate the proposed model on the TACoS dataset.
The results and comparison with 17 representative methods are listed in Table \ref{tab:02}.
For clarity, we mark the best results in bold font.
The TACoS is rather challenging due to the same kitchen background
with some slightly varied cooking objects.
Our framework consistently outperforms these state-of-the-art methods,
w.r.t all metrics with the C3D model.
More specially, our method significantly outperforms the previous
best methods \cite{2DTAN} by 5.25\% on the TACoS.
This remarkable performance demonstrates that utilizing language to guide the moment retrieval is of great benefit to localizing such fine-grained activities.

\subsection{Ablation Studies}
In this subsection, we perform detailed ablation studies on the Charades-STA and TACoS
to demonstrate the effects of the proposed framework components.
The results are shown in Table \ref{tab:03} and Table \ref{tab:04}, respectively.

\textbf{The effect of early modulation unit.}
Inspired by the outstanding performance of 2D-TAN \cite{2DTAN}, we adopt it as a strong baseline.
For fair comparisons, we adopt the same textual features and
hyper-parameters settings with 2D-TAN \cite{2DTAN}.
In order to validate the effectiveness of early fusion (in Equation \ref{equ:02}),
we only modulate the visual processing via the sentence embedding in the early stage.
From the experimental results, we find that in both benchmarks,
early modulation unit gains higher retrieval accuracy,
demonstrating that the proposed early modulation unit can further capture the semantic of sentence query.

We also aim to figure out the question on where should we insert the early module unit.
We first add the early modulation units after the convolutional layers but it leads to degradation of performance.
On the contrary, when we equip the fully-connected layers with this unit, we obtain satisfactory results.
The reason may be that the convolutional blocks usually capture the low-level features
and fully-connected layers always contain more semantic information described in the sentence query. 
Besides, we also adopt the representative modulation method \cite{perez2018film} for moment retrieval. 
However, it obtains slight improvement (less than 1\%).

\textbf{The effect of late guidance module.}
To investigate the effectiveness of the late guidance module (in Equation \ref{equ:03}),
we equip 2D-TAN \cite{2DTAN} with this module.
For fair comparisons, we replace the visual features from 2D-TAN \cite{2DTAN} with the features extracted by this VGG model \footnote{https://www.dropbox.com/s/p457h2ifi6v1qdz/twostream\_rgb.pth.tar}
since we adopt this model for the early modulation unit.
This substitute brings a slight improvement from 40.94\% to 42.34\%.

Based on the assumption that the recall values under higher IoU are more stable because of the dataset biases,
we list the corresponding experimental results for these two modules in terms of Rank1@IOU0.5.
From both Tables, we can observe that the late guidance module dramatically
outperforms the strong baseline on both datasets.
According to Table \ref{tab:04}, the late guidance module improves the
retrieval performance by 4.3\% on the challenging TACoS dataset.

We also verify the late guidance module on the YouMakeup dataset \cite{wang2019youmakeup},
which is a large-scale and multimodal benchmark for
fine-grained semantic comprehension in the makeup domain.
Specifically, the late guidance module boosts the  Rank1@IoU0.5 from 39.02\% to 41.17\% on the validation set.
Moreover, our method also outperforms one of the state-of-the-art models,
SCDM \cite{yuan2019semantic} (34.39\%) and SCDM trained with more supervision (37.33\%)
with the same I3D \cite{carreira2017quo} visual features\footnote{https://github.com/AIM3-RUC/Youmakeup\_Baseline}.

\renewcommand\arraystretch{1.1}
\begin{table}[t]\small
\begin{center}
\newcommand{\tabincell}[2]{\begin{tabular}{@{}#1@{}}#2\end{tabular}}
\centering \caption{Ablation Studies on Charades-STA dataset.}
\label{tab:03}
\vspace{1mm}
\begin{tabular}{ccccccc}
  \toprule
          Methods   &\tabincell{c}{Rank1@IOU0.5}  &\tabincell{c}{$\Delta$}\\
  \midrule
      Baseline         &42.34 &0.0\\
  \midrule
      Baseline with early modulation               &44.17 &1.83\\
  \midrule
      Baseline with late guidance         &46.08 &3.74\\
  \midrule
      Ours             &48.15 &5.81\\
    \bottomrule
\end{tabular}
\end{center}
\end{table}

\renewcommand\arraystretch{1.1}
\begin{table}[t]\small
\begin{center}
\newcommand{\tabincell}[2]{\begin{tabular}{@{}#1@{}}#2\end{tabular}}
\centering \caption{Ablation Studies on TACoS dataset.}
\label{tab:04}
\vspace{1mm}
\begin{tabular}{ccccccc}
  \toprule
          Methods   &\tabincell{c}{Rank1@IOU0.5}  &\tabincell{c}{$\Delta$}\\
  \midrule
      Baseline         &25.32  &0.0\\
  \midrule
      Baseline with early modulation               &27.63 &2.31\\
  \midrule
      Baseline with late guidance         &29.57 &4.25\\
  \midrule
      Ours            &30.57 &5.25\\
    \bottomrule
\end{tabular}
\end{center}
\end{table}

\textbf{The effect of both modules.}
According to Table \ref{tab:03},
combining both early modulation unit and late guidance module can obtain substantial performance gains from 42.34\% to 48.15\% on Charades-STA,
which improves the single module and demonstrates the effectiveness of building both modules to bridge the vision and language domains.
Besides, we see that late guidance modules achieve 3.74\% improvements over the
baseline and early modulation units perform 1.83\% better than the baseline.
Therefore, it seems that the late guidance module has a greater effect on performance than early modulation units.

\textbf{The effect of hyper-parameters.}
To investigate the impact of hyper-parameters and acquire the best,
we conduct several experiments based on the visual features from 2D-TAN \cite{2DTAN} using the different number of late guidance modules.
We note that more modules can achieve higher performance.
Besides, the performance saturates when the number of modules greater than $6$.
Thus, we set the number of late guidance modules to $6$ as the default setting.
Since the early modulation units require to be equipped around the fully-connected layers, we adopt two early modulation units.

\section{Conclusion}
\label{concl}
In this paper, we study the problem of moment localization with natural language and propose Language Guided Networks (LGN)
to leverage the sentence embedding to guide the whole process of moment retrieval.
The core idea is to modulate the visual processing conditioned by the language in the early stage,
and linearly transform the outputs of localization networks with the guidance of sentence embedding in the late stage.
As a result, our method is capable of learning discriminative visual features
and tightly integrates cross-modal features.
Besides, our model is simple in design and achieves performance superior to the state-of-the-art methods on the two popular datasets.
In the future, 
we would like to extend our method to other vision and language tasks,
such as video question answering, video captioning, etc.
\begin{small}
\bibliography{reference}
\end{small}

\end{document}